\documentclass[final]{elsarticle}

\usepackage{graphicx}
\usepackage{amsmath}
\usepackage{amssymb}
\usepackage{algorithmic}
\usepackage{algorithm}
\usepackage{subfigure}
\usepackage{array}
\usepackage{multirow,tabularx}

\newcommand{\bfx}{{\textbf{x}}}

\newcommand{\bfw}{{\textbf{w}}}

\journal{Engineering Applications of Artificial Intelligence}

\begin{document}

\begin{frontmatter}

\title{Maximum mutual information regularized classification}

\author[KAUST]{Jim Jing-Yan Wang}
\ead{jimjywang@gmail.com}

\author[OSU]{Yi Wang}
\ead{wayi@cse.ohio-state.edu}

\author[HMU]{Shiguang Zhao}
\ead{guangsz@hotmail.com}

\author[KAUST]{Xin Gao \corref{cor1}}
\ead{xin.gao@kaust.edu.sa}

\address[KAUST]{Computer, Electrical and Mathematical Sciences and Engineering Division, King Abdullah University of Science and Technology (KAUST), Thuwal 23955-6900, Saudi Arabia}

\address[OSU]{Department of Computer Science and Engineering,
The Ohio State University, Columbus OH 43210, USA}

\address[HMU]{Department of Neurosurgery, The First Affiliated Hospital of Harbin Medical University, Harbin, Heilongjiang 150001, P.R. China}

\cortext[cor1]{To whom all correspondence should be addressed. Tel:  +966-12-8080323.}

\begin{abstract}
In this paper,  a novel pattern classification approach is proposed by regularizing the classifier learning to
maximize  mutual information between the classification response and the true class label.
We argue that, with the learned classifier, the uncertainty of the true class label of a data sample should be reduced by knowing its classification response as much as possible.
The reduced uncertainty is measured by the mutual information between the classification response and the true class label. To this end, when learning a linear classifier, we propose to maximize the mutual information between classification responses and true class labels of training samples, besides minimizing the classification error and reducing the classifier complexity.
An objective function is constructed by modeling mutual information with entropy estimation, and it is optimized by a gradient descend method in an iterative algorithm.
Experiments on two real world pattern classification problems show the significant improvements achieved by maximum mutual information regularization.
\end{abstract}


\end{frontmatter}

\section{Introduction}

The pattern classification problem is a problem of assigning a discrete class label to a given data sample
represented by its feature vector \cite{Cai2014258,Ojala2002971,sun2014mobile,li2014doc,li2014graph}.
It has many applications in various fields, including bioinforamtics
\cite{alipanahi2009picky,wang2013non,wang2012multiple,wang2012prodis,liu2012wavpeak},
biometrics verification \cite{Wang2009IS21,roy2011towards,tafazzoli2010model},
computer networks \cite{yang2014location,CNS:Evasion,xu2013cross}, and
computer vision \cite{6607606,wang2014sparse,wang2013joint,zhou2010region}.
For example, in the face recognition problem, given a face image,
the target of pattern classification is to assign it to a person who has been
registered in a database \cite{JonathonPhillips20001090,Zhao2003399}.
This problem is usually composed of two different components ---
feature extraction  \cite{sun2012unsupervised,Zhou2013190,wang2013beyond,al2014supervised,wang2013feature,wang2013discriminative,wang2012adaptive,wang2013multiple} and classification \cite{zhou2013adaptive,Subbulakshmi2013308}.
Feature extraction refers to the procedure of extracting an effective and discriminant feature vector from a data sample, so that
different samples of different classes could be separated easily.
This procedure is usually highly domain-specific.
For example, for the face recognition problem, the visual feature should be extracted using
some image processing technologies, whereas for the problem of predicting zinc-binding sites from protein sequences,
the biological features should be extracted using some biological knowledge \cite{Chen20132213}.
In terms of feature extraction of this paper, it is highly inspired by a hierarchical Bayesian inference algorithm proposed in [24]. This new method created in \cite{sun2012unsupervised} has advanced the ground-truth feature extraction field and has provided a more optimal method for this procedure.
On the other hand, different from feature extraction, classification is a much more general
problem.
We usually design a class label prediction function as a classifier for this purpose.
To learn the parameter of a classifier function,
we  usually try to minimize the classification error of the training samples in a training set
and simultaneously reduce the complexity of the classifier. For example, the most popular classifier is support vector machine (SVM), which minimizes the hinge losses to reduce the classification error, and at the same time minimizes the $\ell_2$ norm of the classifier parameters to reduce the complexity.
In this paper, we focus on the classification aspect.

Mutual information \cite{LCS2014,Battiti1994537} is defined as the information
shared between two sets of variables.
It has {been} used as a criterion of feature extraction for pattern classification problems \cite{Sun2014763}.
However, surprisingly, it has never been directly explored in the problem of classifier learning.
{Actually, mutual information has a strong relation to Kullback-Leibler divergence, and there are many works using KL-divergence for classifiers \cite{moreno2003kullback,liu2003kullback}.
Moreno et al. \cite{moreno2003kullback} proposed a novel kernel function for support vector classification based on
Kullback-Leibler divergence, while Liu and Shum \cite{liu2003kullback} proposed to learn the most discriminating feature that maximizes the Kullback-Leibler divergence for the Adaboost classifier. However, both these methods do not use the KL-divergence based criterion to learn parameters of linear classifiers.}
To bridge this gap, in this paper, for the first time, we try to investigate using mutual information as a criterion of classifier learning.
We  propose to learn a classifier by  maximizing the mutual information $I(f;y)$ between the
classification response variable $f$ and the true class label variable $y$.
The classification response variable $f$ is a function of classifier parameters and data samples.
The insight is that mutual information is defined as the
information shared between $f$ and $y$.
From the viewpoint of information theory, if the two variables are not mutually independent,
and one variable is known, it usually  reduces the
uncertainty about the other one.
Then mutual information is used to measure how much uncertainty is reduced in this case.
To illuminate how the  mutual information can be used to measure the
classification accuracy, we consider the two extreme cases:
\begin{itemize}
\item
On one hand, if the classification response variable $f$ of a data sample is randomly given, and it is
independent of its true class label $y$, then
knowing $f$ does not give any
information about $y$ and vice versa,
and the mutual information between them could be zero, i.e.,  $I(f;y)=0$.

\item
On the other hand, if $f$ is given so that $y$ and $f$ are identical, knowing $f$ can help determine the value of $y$ exactly as well as reduce all the uncertainty about $y$.
This is the ideal case of classification, and
knowing $f$ can reduce all the uncertainty about $y$.
In this case, the
mutual information is defined as the uncertainty contained in $f$ (or $y$) alone,
which is measured by the entropy of $f$ or $y$, denoted by $H(f)$ or $H(y)$ respectively, where $H(\cdot)$ is the entropy of a variable. Since f and y are identical, we can have $I(f;y)=H(f)=H(y)$.
\end{itemize}
Naturally, we hope that the classification response $f$ can predict the
true class label $y$ as accurately as possible,
and knowing $f$ can reduce the uncertainty about $y$ as much as possible.
Thus, we propose to maximize the
mutual information between $f$ and $y$ with regard to the parameters of a classifier. To this end, we proposed a mutual information regularization term for the learning of classifier parameters. An objective function is constructed
by combining the mutual information regularization term, a classification error term and a classifier complexity term.
The classifier parameter is learned by optimizing the objective function with a gradient descend method in an iterative algorithm.

The rest parts of this paper are organized as follows: in Section \ref{sec:met}, we introduce the proposed classifier learning method. The experiment results are presented in section \ref{sec:exp}. In section \ref{sec:con} the paper is concluded.

\section{Proposed Method}
\label{sec:met}

In this section, we introduce the proposed classifier learning algorithm to maximize the mutual information between the classification response and the true class label.

\subsection{Problem Formulation}

We suppose that we have a training set denoted as
$X=\{\bfx_i\}_{i=1}^n$,
where $\bfx_i\in \mathbb{R}^d$ is the $d$-dimensional feature vector for the $i$-th training sample,
and $n$ is the number of training samples.
The class label set for the training samples is denoted as
$Y=\{y_i\}_{i=1}^n$,
where $y_i\in \{+1,-1\}$ is the class label of the $i$-th sample.
To learn a classifier to predict the class label of a given sample with its feature vector $\bfx$,
we design a linear function as a classifier,

\begin{equation}
\begin{aligned}
g(\bfx;\bfw)=sign(f)=sign(\bfw^\top \bfx),
\end{aligned}
\end{equation}
where $\bfw$ is the classifier parameter vector, $f=\bfw^\top \bfx$ is the classification response of $\bfx$ given
the classifier parameter $\bfw$, and $sign(\cdot)$ is the signum function which transfers the classification response to the final binary classification result.
We also denote the classification response set of the training samples as
$F=\{f_i\}_{i=1}^n$
where $f_i=\bfw^\top \bfx_i \in \mathbb{R}$
is the classification response of the $i$-th training sample.
To learn the optimal classification parameter $\bfw$ for the classification problem, we consider the following three problems:


\subsubsection{Classification Loss Minimization}

To learn the optimal classification parameter $\bfw$, we hope the classification
response $f$ of a data sample $\bfx$ obtained with the learned $\bfw$ can predict
its true class label $y$ as accurately as possible.
To measure the prediction error, we use a  loss function  to compare a classification response against its corresponding true class label.
Given the classifier parameter $\bfw$, the loss function of the $i$-th  training  sample $\bfx_i$ with its classification response
$f_i=\bfw^\top \bfx_i$ and true class label $y_i$ is denoted as $L(f_i,y_i;\bfw )$.
There are a few different loss functions which could be considered.

\begin{description}
\item [Hinge  Loss] is used by the SVM classifier \cite{Wu2007974,Yildiz2013153,Bach201332}, and it is defined as

\begin{equation}
\begin{aligned}
L(f_i,y_i;\bfw )
=
\max
(0,1-y_i f_i)
=
\tau_i \times
(1-y_i\bfw^\top\bfx_i),
\end{aligned}
\end{equation}
where $\tau_i$ is defined as

\begin{equation}
\label{equ:tau}
\begin{aligned}
\tau_i
=
\left\{\begin{matrix}
1 ,& if ~ y_i \bfw^\top\bfx_i \leq 1\\
0 ,& otherwise.
\end{matrix}\right.
\end{aligned}
\end{equation}

\item [Squared  Loss] is usually used by regression problems \cite{Wang2013632,luo2012regression,luo2012based}, and it is defined as

\begin{equation}
\begin{aligned}
L(f_i,y_i;\bfw )
&=
(1- y_i f_i)^2
=(1- y_i \bfw^\top \bfx_i)^2.
\end{aligned}
\end{equation}

\item [Logistic Loss] is defined as follows, and it is also popular in regression problems \cite{Park20083709},

\begin{equation}
\begin{aligned}
L(f_i,y_i;\bfw )
=
\log \left [1+  \exp(-y_i f_i)\right]
=
\log \left [1+  \exp(-y_i \bfw^\top \bfx_i)\right].
\end{aligned}
\end{equation}

\item [Exponential loss] is anther popular loss function which could be used by both classification and regression problems \cite{Wang2013632}, which is defined as

\begin{equation}
\begin{aligned}
L(f_i,y_i;\bfw )
=
\exp(-y_i f_i)=
\exp(-y_i \bfw^\top \bfx_i).
\end{aligned}
\end{equation}

\end{description}
Obviously, to learn an optimal classifier, the  average loss of all the training samples should be minimized with regard to $\bfw$.
Thus the following optimization problem is obtained by applying a loss functions to all training samples,

\begin{equation}
\label{equ:ob2}
\begin{aligned}
\min_{\bfw}~
\frac{1}{n}
\sum_{i=1}^n L(f_i,y_i;\bfw).
\end{aligned}
\end{equation}

\subsubsection{Classifier Complexity Reduction}

To reduce the complexity of the classifier to prevent the over-fitting problem, we also regularize the classifier parameter by a $\ell_2$ norm term as

\begin{equation}
\label{equ:ob3}
\begin{aligned}
\min_{\bfw}~
\frac{1}{2}\|\bfw\|^2_2.
\end{aligned}
\end{equation}

\subsubsection{Mutual Information Maximization}

We  also propose to learn the classifier by  maximizing the mutual information $I(f;y)$ between the
classification response variables $f\in F$ and
the true class label variables $y\in Y$.
The mutual information between two variables $f\in F$
and $y\in Y$
is defined as

\begin{equation}
\begin{aligned}
I(f;y)=
H(f) - H(f|y),
\end{aligned}
\end{equation}
where $H(f)$ is  the marginal entropy of $f$,
which is used to measure
the uncertainty about $f$, and $ H(f|y)$ is the entropy
of $f$ conditional on $y$, which is used as
the measure  of uncertainty of $f$ when $y$ is given.
To use the mutual information as  a criterion
to learn the classifier parameters, we first need to estimate $H(f)$
and $H(f|y)$.

\begin{description}
\item[Estimation of $H(f)$]
We use the training samples to estimate $H(f)$,
and according to the definition of entropy, we have

\begin{equation}
\label{equ:H_f}
\begin{aligned}
H(f)
&=-
\int p(f) \log p(f) d f\\
&\approx
-
\sum_{i=1}^n p(f_i) \log p(f_i),
\end{aligned}
\end{equation}
where $p(f)$ is the probability density  of $f$.
It could be seen that the entropy of $f$ is the expectation of
$-\log p(f)$ \cite{Bekenstein19732333}.
The non-parametric kernel density estimation (KDE)  \cite{Elgammal20021151}
is used to estimate the probability density function $p(f)$,

\begin{equation}
\label{equ:p_f}
\begin{aligned}
p(f)
&=
\frac{1}{n}
\sum_{j=1}^n
K
\left(f-f_j;\sigma
\right),
\end{aligned}
\end{equation}
where
$
K
\left(z;\sigma
\right)
=
exp
\left (
-\frac{z^2}{2 \sigma^2}
\right )
$
is the Gaussian kernel function \cite{Zhong20132045} and $\sigma$ is the bandwidth parameter \cite{Elgammal20021151}.

\item[Estimation of $H(f|y)$]
We also use the training samples to estimate $H(f|y)$,
and according to its definition, we have

\begin{equation}
\label{equ:H_fy}
\begin{aligned}
H(f|y)
&=
\sum_{c\in \{+1,-1\}} p(c)H(f|y=c),
\end{aligned}
\end{equation}
where
$
p(c)=\frac{n_c}{n}
$
is the probability density of class label $c$, $n_c = \# \{\bfx_i\in X|y_i=c\}$ is the number of samples with the class label equal to $c$, and

\begin{equation}
\label{equ:H_f_y}
\begin{aligned}
H(f|y=c)
&=-
\int p(f|y=c) \log p(f|y=c) d f\\
&\approx
-
\sum_{i:y_i=c}
p(f_i|y=c)
\log p(f_i|y=c)
\end{aligned}
\end{equation}
is the conditional entropy of $f$ given
the class label $y=c$ \cite{Carvalho20132716,Porta199871,Wang2002759}.
We also use the KDE to estimate the conditional
probability density function

\begin{equation}
\begin{aligned}
p(f|y=c)=
\frac{1}{n_c}
\sum_{i:y_i=c}
K
\left(f-f_i,\sigma
\right)
\end{aligned}
\end{equation}
Substituting it to (\ref{equ:H_fy}), we have the estimated $H(f|y)$,

\begin{equation}
\begin{aligned}
H(f|y)
&\approx
-
\sum_{c\in \{+1,-1\}} \frac{n_c}{n}
\left (
\sum_{i:y_i=c}
p(f_i|y=c)
\log
p(f_i|y=c)
\right )
\end{aligned}
\end{equation}

\end{description}
With the estimated entropy $H(f)$ and the conditional entropy $H(f|y)$, the
mutual information between the variable $f$ and $y$ could be rewritten as
the function of parameter $\bfw$ by substituting $f_i=\bfw^\top \bfx_i$,

\begin{equation}
\begin{aligned}
\widetilde{I}(f,y;\bfw)
=&H(f)-H(f|y)\\
=&
-
\sum_{i=1}^n p(f_i) \log p(f_i)\\
&
+
\sum_{c\in \{+1,-1\}} \frac{n_c}{n}
\left (
\sum_{i:y_i=c}
p(f_i|y=c)
\log
p(f_i|y=c)
\right )\\
=&
-
\sum_{i=1}^n p(\bfw^\top \bfx_i) \log
p(\bfw^\top \bfx_i)\\
&
+
\sum_{c\in \{+1,-1\}} \frac{n_c}{n}
\left (
\sum_{i:y_i=c}
p(\bfw^\top \bfx_i|y=c)
\log
p(\bfw^\top \bfx_i|y=c)
\right )
\end{aligned}
\end{equation}
To learn the classifier parameter $\bfw$, we  maximize the mutual information with regard to $\bfw$,

\begin{equation}
\label{equ:ob1}
\begin{aligned}
\max_{\bfw}~
\widetilde{I}(f,y;\bfw)
\end{aligned}
\end{equation}

{\textbf{Remark}: It should be noted that similar to our method, the algorithm proposed in \cite{liu2003kullback} maximizes KL-divergence between the class PDF, $p(f|y=c)$, and the total PDF, $p(f)$, therefore, \cite{liu2003kullback} has relation to method in Kullback-Leibler boosting. However, different from our method, it uses KL-divergence as a criterion to select the most discriminating features, whereas our method uses mutual information as a criterion to learn the classifier parameter.}

\subsubsection{Overall Optimization Problem}

By combining the optimization problems proposed in (\ref{equ:ob2}), (\ref{equ:ob3}) and  (\ref{equ:ob1}),
the optimization problem for the proposed classifier parameter learning method is
obtained as

\begin{equation}
\label{equ:og4}
\begin{aligned}
\min_{\bfw}~
\left \{ O(\bfw)=
\frac{1}{n} \sum_{i=1}^n L(f_i,y_i;\bfw)
+
\alpha
\frac{1}{2}\|\bfw\|^2_2
-\beta\widetilde{I}(f,y;\bfw)
\right \},
\end{aligned}
\end{equation}
where $\alpha$ and $\beta$ are tradeoff parameters.
In the objective function, there are three terms.
The first one is optimized so that the
prediction error is minimized,
the second term is used to contral the complexity of the classifier,
and the last term is introduced so that the mutual information between the classification response
and the true class label can be maximized.

\subsection{Optimization}

Direct optimization to  (\ref{equ:og4}) is difficult.
Instead of seeking a closed-form solution, we try to optimize it
using gradient descent method in an iterative algorithm \cite{dempster1977maximum}.
In each iteration, we employ the gradient descent method to update $\bfw$.
According to the optimization theory,
if $Q(\bfw)$ is
defined and differentiable in a neighborhood of a point $\bfw^{old}$, then
$Q(\bfw)$ decreases faster if $\bfw$ goes from $\bfw^{old}$ in the direction of the
negative gradient of $Q(\bfw)$ at $\bfw^{old}$,
$-   \nabla Q(\bfw) |_{\bfw=\bfw^{old}}$. Thus the new $\bfw$ is obtained by

\begin{equation}
\label{equ:w}
\begin{aligned}
\bfw^{new}\leftarrow
\bfw^{old} -  \eta \nabla Q(\bfw) |_{\bfw=\bfw^{old}},
\end{aligned}
\end{equation}
where $\eta$ is the descent step.

The key step is to compute the gradient of $Q(\bfw)$,  which is calculated as

\begin{equation}
\begin{aligned}
\nabla Q(\bfw) =
\frac{1}{n} \sum_{i=1}^n \nabla L(f_i,y_i;\bfw)
+
\alpha
\bfw
-\beta\nabla\widetilde{I}(f,y;\bfw),
\end{aligned}
\end{equation}
where $\nabla L(f_i,y_i;\bfw)$ and $\nabla\widetilde{I}(f,y;\bfw)$ are the gradient of $L(f_i,y_i;\bfw)$ and $\widetilde{I}(f,y;\bfw)$ respectively. They are given analytically as follows.

\subsubsection{Computation of $\nabla L(f_i,y_i;\bfw)$}

We give the analytical gradients of different definitions of $L(f_i,y_i;\bfw)$  as follows:

\begin{description}
\item [Hinge  Loss] is not a smooth function, but we can first update $\tau_i$ using previous $\bfw$ as in (\ref{equ:tau}), and then fix it when we derivate $\nabla L(f_i,y_i;\bfw )$,

\begin{equation}
\begin{aligned}
\nabla L(f_i,y_i;\bfw )
=
-\tau_i \times
(y_i\bfx_i).
\end{aligned}
\end{equation}

\item [Squared  Loss] is a smooth function, and its gradient is

\begin{equation}
\begin{aligned}
\nabla L(f_i,y_i;\bfw )
= -  (1- y_i \bfw^\top \bfx_i) \times ( y_i   \bfx_i).
\end{aligned}
\end{equation}

\item [Logistic Loss] is also smooth with its gradient as

\begin{equation}
\begin{aligned}
\nabla L(f_i,y_i;\bfw )
= -
\frac{\exp(-y_i \bfw^\top \bfx_i)}{ 1+  \exp(-y_i \bfw^\top \bfx_i)}
\times
(y_i  \bfx_i).
\end{aligned}
\end{equation}

\item [Exponential loss] is also smooth, and its gradient can be obtained as

\begin{equation}
\begin{aligned}
\nabla L(f_i,y_i;\bfw )
= -
\exp(-y_i \bfw^\top \bfx_i)
\times
(y_i  \bfx_i).
\end{aligned}
\end{equation}

\end{description}

\subsubsection{Computation of $\nabla\widetilde{I}(f,y;\bfw)$}

The gradient of $\widetilde{I}(f,y;\bfw)$ is computed as

\begin{equation}
\label{equ:I}
\begin{aligned}
\nabla \widetilde{I}(f,y;\bfw)
=&
-\sum_{i=1}^n
\left( \nabla p(\bfw^\top \bfx_i) \log p(\bfw^\top \bfx_i)+
\nabla p(\bfw^\top \bfx_i)
\right )
\\
&+
\sum_{c\in \{+1,-1\}} \frac{n_c}{n}
\left [
\sum_{i:y_i=c}
\left (
\nabla p(\bfw^\top \bfx_i|y=c) \log p(\bfw^\top \bfx_i|y=c)
+
\nabla p(\bfw^\top \bfx_i|y=c)
\right )
\right ]\\
=&-
\sum_{i=1}^n
\left(\log p(\bfw^\top \bfx_i)+ 1
\right ) \nabla p(\bfw^\top \bfx_i)
\\
&+
\sum_{c\in \{+1,-1\}} \frac{n_c}{n}
\left (
\sum_{i:y_i=c}
\left ( \log p(\bfw^\top \bfx_i|y=c) +  1 \right ) \nabla p(\bfw^\top \bfx_i|y=c)
\right ),
\end{aligned}
\end{equation}
where the gradients of $p(\bfw^\top \bfx_i)$ and $p(\bfw^\top \bfx_i|y=c)$ are computed as

\begin{equation}
\begin{aligned}
\nabla p(\bfw^\top \bfx_i)
&=\frac{1}{n \sigma^2}
\sum_{j=1}^n
\exp
\left (
-\frac{(\bfw^\top\bfx_i-\bfw^\top\bfx_j)^2}{2 \sigma^2}
\right )
(\bfw^\top\bfx_i-\bfw^\top\bfx_j)(\bfx_j-\bfx_i),\\
\nabla p(\bfw^\top \bfx_i|y=c)
&=\frac{1}{n_c \sigma^2}
\sum_{j:y_i=c}
\exp
\left (
-\frac{(\bfw^\top\bfx_i-\bfw^\top\bfx_j)^2}{2 \sigma^2}
\right )
(\bfw^\top\bfx_i-\bfw^\top\bfx_j)(\bfx_j-\bfx_i).
\end{aligned}
\end{equation}

\section{Experiments}
\label{sec:exp}

In this section, we evaluate the proposed classification method on two real world pattern classification problems.

\subsection{Experiment I: Zinc-binding Site Prediction}

Zinc is an important element for many biological processes of an organism, and it is closely related to many different  diseases.
Moreover, it is also critical for proteins to play their functional roles \cite{Chen20132213,Kumar20133949,Liu2014171}.
Thus functional annotation of zinc-binding proteins is necessary to  biological process control and disease treatment.
To this end, predicting zinc-binding sites of proteins shows its importance in bioinformatics problems.
In the first experiment, we evaluate the proposed classification method on the problem of predicting zinc-binding sites.

\subsubsection{Data Set and Protocol}

For the purpose of experiment, we collected a set of amino acids of four types, which are CYS, HIS, GLU and ASP (CHED).
These four types are the most common zinc-binding site types, which take up roughly 96\% of the known zinc-binding {sites}.
In the collected data set, there are 1,937 zinc-binding CHEDs and 11,049  non-zinc-binding CHEDs, resulting a data set of 13,986 data samples.
Given a candidate CHED, the problem of zinc-binding site prediction is to predict if it is a zinc-binding site or a non-zinc-binding site.
In this experiment, we treated a  zinc-binding CHED as a positive sample, and a non-zinc-binding CHED as a negative sample.
To extract features from a CHED, we computed the
position specific substitution matrices (PSSM) \cite{Kelley2000499},
the relative weight of gapless real matches to pseudocounts (RW-GRMTP) \cite{menchetti2006improving},
Shannon entropy \cite{Chen20132213},
and composition of $k$-spaced amino acid pairs (CKSAAP) \cite{Zhang2013911},
and concatenated them to form a feature vector for each data sample.
{Please note that the value of each feature was scaled to the range between -1 and 1, so that the performance does not depend on the selection of scaling.}

To conduct the experiment, we used the 10-fold cross validation protocol \cite{Burman1989503}. The entire data set was split into ten non-overlapping folds,  and each fold was used as a test set in turn, while the remaining nine folds were combined and used as a training set.
The proposed algorithm was performed to the training set to learn a classifier from the feature vectors of the training samples,
and then the learned classifier was used to predict the class labels of the test samples.
Please note that the tradeoff parameters of the proposed algorithm was tuned within the training set.
{The averaged value of the hyper-parameters $\alpha$ and $\beta$ are 5.8 and 44.8.
The parameter $\sigma$ was computed as $\sigma = \varsigma \times dist$, where $dist$ was the median value of distances between pairs of training samples, and the averaged value of $\varsigma$ was 0.451.}
The classification performance was measured by comparing the predicted labels against the true labels.
The receiver operating characteristic (ROC) and recall-precision curves were used as performance metrics.
The ROC curve was obtained by plotting true positive rates (TPRs) against false positive rates (FPRs), while recall-precision curve was obtained by plotting precision against recall values.
TPR, FPR, recall, and precision are defined as

\begin{equation}
\begin{aligned}
&TPR=\frac{TP}{TP+FN}, FPR=\frac{FP}{FP+TN},\\
&recall=\frac{TP}{TP+FN}, precision= \frac{TP}{TP+FP},
\end{aligned}
\end{equation}
where $TP$, $FP$, $FN$ and $TN$ represent the number of true positives, false
positives, false negatives and true negatives, respectively.
Moreover, area under ROC curve (AUC) \cite{Zou20135077} was used as a single performance measure.
A good classifier should achieve a ROC curve close to the top left corner of the figure,
a recall-precision curve close to the top right corner, and also a high AUC value.

\subsubsection{Resutls}

\begin{figure*}
\centering
\subfigure[Hinge Loss]{
\includegraphics[width=0.70\textwidth]{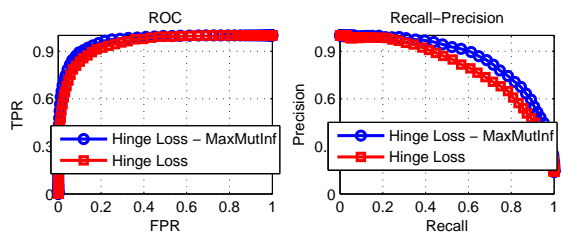}}
\subfigure[Squared Loss]{
\includegraphics[width=0.70\textwidth]{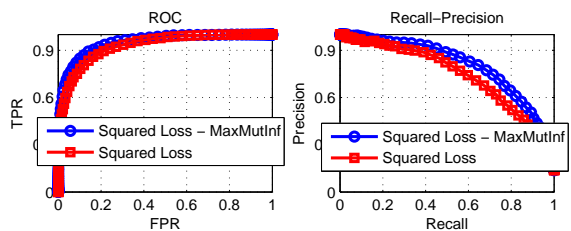}}
\subfigure[Logistic Loss]{
\includegraphics[width=0.70\textwidth]{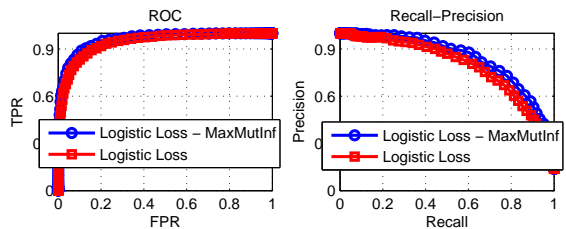}}
\subfigure[Exponential Loss]{
\includegraphics[width=0.70\textwidth]{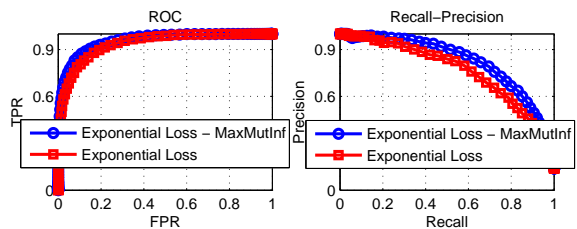}}
\caption{ROC and recall curves of experiments on zinc-binding site prediction.
{``Loss'' stands for combination of classification loss and $\ell_2$-regularization, and ``Loss - MaxMutInf'' stands for combination of classification loss, $\ell_2$ and maximum mutual information-regularization.}}
\label{fig:ZincROC}
\end{figure*}

In this experiment, we compared the proposed mutual information regularized classifier against
the original loss functions based classifier without mutual information regularization,
so that the improvement achieved by maximum mutual information regularization could be verified.
The four different loss functions listed in Section \ref{sec:met} were considered, and the corresponding classifiers were evaluated here.
The ROC and recall-precision curves of four loss functions based classification methods are given in Fig. \ref{fig:ZincROC}.
The proposed  maximum mutual information regularized method is denoted as ``MaxMutInf" after a loss function title in
the figure.
It turns out that maximum mutual information regularization improves all the four loss functions based
classification methods significantly.
Although various loss functions achieved different performances, all of them could be boosted by reducing
the uncertainty about true class labels,  which could be measured by the mutual information between class labels and classification responses. Therefore, the results show that maximizing mutual information is highly effective in reducing uncertainty of true class labels, and hence it can significantly improve the quality of classification.

\begin{figure}[!h]
\centering
\includegraphics[width=0.7\textwidth]{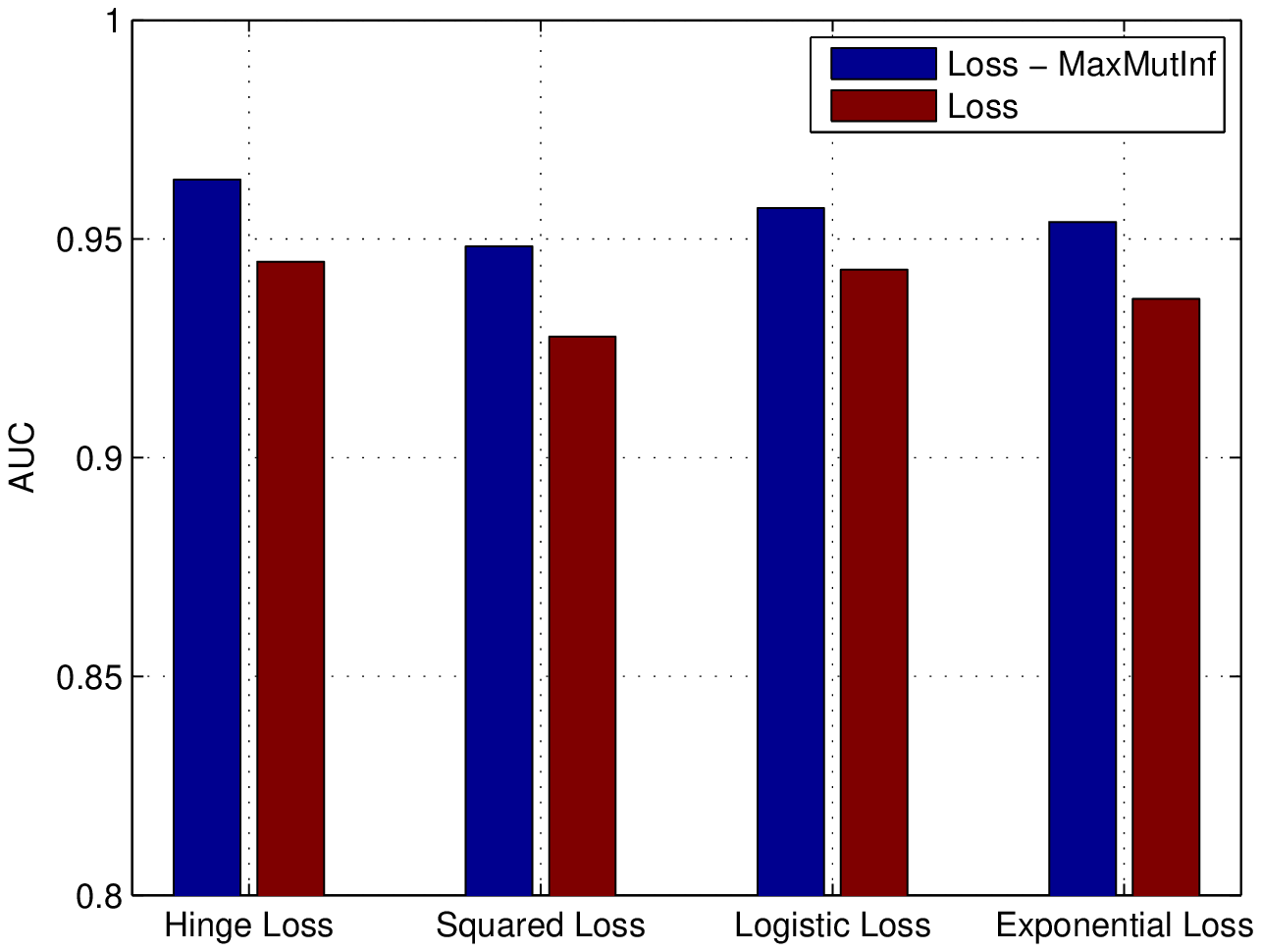}\\
\caption{AUC values of experiments on zinc-binding site prediction.
{``Loss'' stands for combination of classification loss and $\ell_2$-regularization, and ``Loss - MaxMutInf'' stands for combination of classification loss, $\ell_2$ and maximum mutual information-regularization.}}
\label{fig:ZincAUC}
\end{figure}

Moreover, we also plotted AUCs of different methods in Fig. \ref{fig:ZincAUC}.
Again, we observe that maximum mutual information regularization improves different loss functions based
classifiers. We also can see that among these four loss functions, hinge loss achieves  the highest
AUC values, while squared loss achieves the lowest.
The AUC value of classifiers regularized by both hinge loss and mutual information is 0.9635,
while that of squared loss and mutual information is even lower than 0.95.
The performances of logistic and exponential loss functions are similar, and they are between the performances of
hinge loss and squared loss.

\begin{figure}[!htb]
\centering
\includegraphics[width=0.7\textwidth]{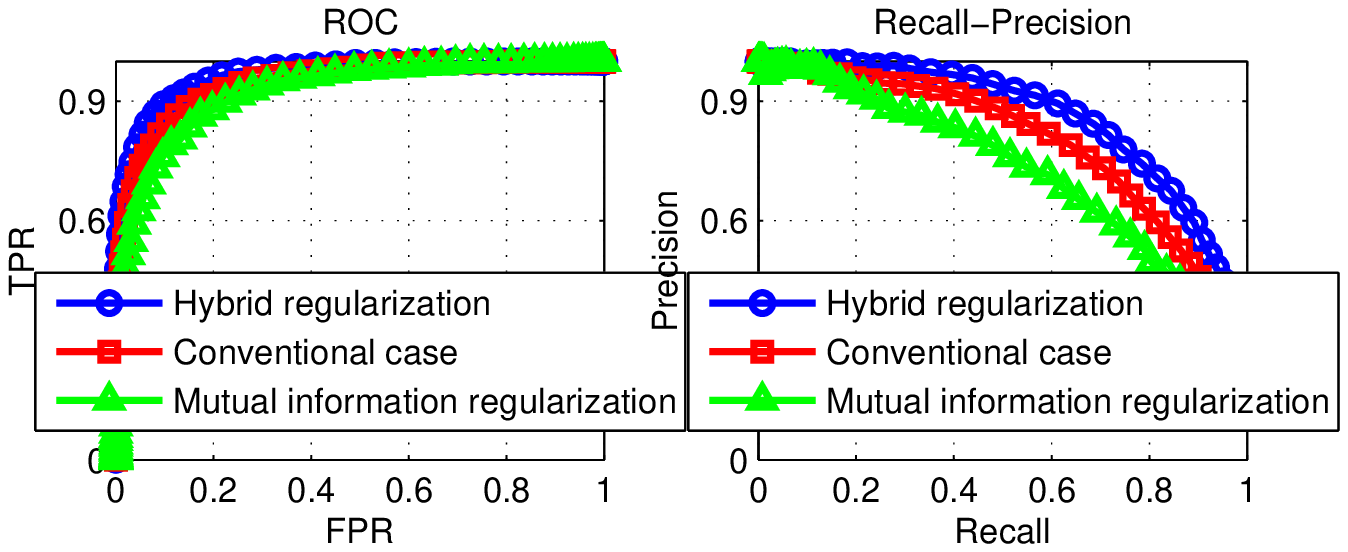}\\
\caption{Comparison results of three regularization cases on zinc-binding site prediction.}
\label{fig:FigZincCompare}
\end{figure}

{Since the mutual information is used as a new regularization technique, we are also interested in how the proposed regularization alone works. We therefore compared the following three cases.
\begin{enumerate}
\item \textbf{Conventional case} which only uses the classification loss regularization. This case is corresponding to setting $\beta=0$ in (\ref{equ:og4}). In this case, we only used the hinge loss since it has been shown that this loss function obtains better accuracy than other loss functions.
\item \textbf{Mutual information regularization case} which is corresponding to the problem in (\ref{equ:og4}) when the first term is ignored.
\item   \textbf{Hybrid regularization case} which is the proposed framework which combines the classification loss minimization and mutual information regularization.
\end{enumerate}
The comparison results are given in Fig. \ref{fig:FigZincCompare}. It can be seen that the conventional case which only uses the hinge loss function achieved better results than the method with only mutual information regularization, and the hybrid regularization achieved the best results. This means mutual information regularization cannot obtain good performance by itself and should be used with traditional loss functions.}

\subsection{Experiment II: HEp-2 Cell Image Classification}

Antinuclear Autoantibodies (ANA) test is a technology used to determine whether a human immune system
is creating antibodies to fight against infections.
ANA is usually done by a specific fluorescence pattern of HEp-2 cell images \cite{20134516948275}.
Recently, there is a great need for computer based HEp-3 cell image classification, since manual classification is time-consuming and not accurate enough.
In the second experiment, we will evaluate the performance of the proposed classifier on the problem of classifying
HEp-2 cell images.

\subsubsection{Data Set and Protocol}

\begin{figure}
\centering
\includegraphics[width=0.7\textwidth]{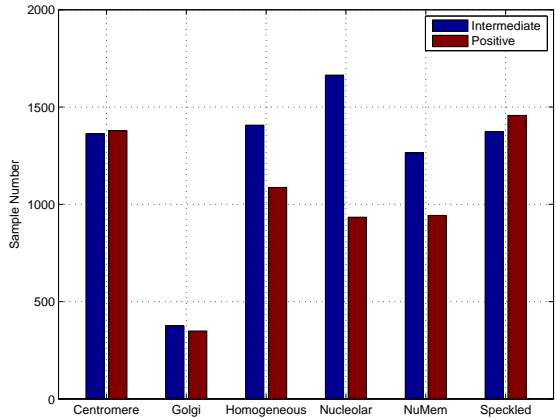}\\
\caption{Number of samples of different classes of the HEp-2 cell image data set.}
\label{fig:HEp2Data}
\end{figure}

In this experiment, we used the database of HEp-2 cell images of the ICIP 2014 competition of cell classification by fluorescent image analysis \cite{bworld}.
In this data set, there are 13,596 cell images, and they belong to six cell classes, which are namely Centromere, Golgi, Homogeneous,
Nucleolar, NuclearMembrane, and Speckled.
Each cell image is segmented by a mask image showing the boundary of the cell.
Moreover, the entire data set is composed of two groups of different tensity types, which are Intermediate and Positive.
Overall, the intermediate group outnumbers the positive group, with an exception that, for the cases of Centromere and Speckled, the latter marginally outnumbers the former.
The number of images in different classes of two groups are given in Fig. \ref{fig:HEp2Data}.
To present each image for the classification problem, we extracted shape and texture features and concatenated them to form a visual  feature vector \cite{20134516948275}.

Experiments were conducted in two groups respectively. We also adopted the 10-fold cross validation for the experiment. To handle the problem of multiple class problem, we used the one-against-all strategy. Each class was treated as a positive class in turn, while all remaining five classes were combined to form a negative class. A classifier was learned for each class to discriminate it from other classes.
A test sample was assigned to a class with the largest classification response.
The classification accuracy was used as a classification performance metric.

\subsubsection{Results}

\begin{figure}
\centering
\subfigure[Hinge Loss]{
\includegraphics[width=0.8\textwidth]{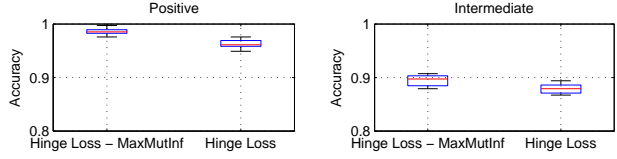}}\\
\subfigure[Squared Loss]{
\includegraphics[width=0.8\textwidth]{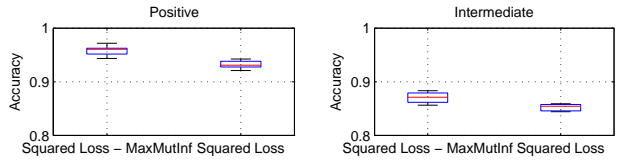}}\\
\subfigure[Logistic Loss]{
\includegraphics[width=0.8\textwidth]{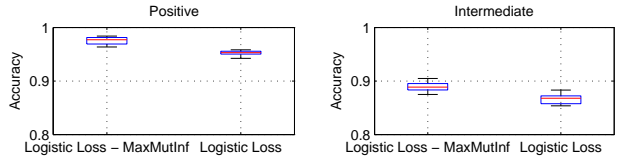}}\\
\subfigure[Exponential Loss]{
\includegraphics[width=0.8\textwidth]{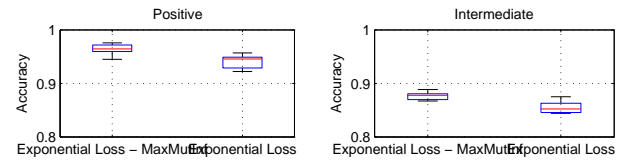}}\\
\caption{Experiment results on HEp-2 cell image data set.
{Please note that ``Loss'' stands for combination of classification loss and $\ell_2$-regularization, and ``Loss - MaxMutInf'' stands for combination of classification loss, $\ell_2$ and maximum mutual information-regularization.}}
\label{fig:HEp2Result}
\end{figure}

The boxplots of accuracies of the 10-fold cross validation on the two groups of HEp-2 cell image data set
are given in Fig. \ref{fig:HEp2Result}. From this figure, it could be observed that for both two groups of data sets, the proposed regularization method can improve the classification performances significantly, despite of the variety of loss functions.
It can also be seen that the performances on the second group (Positive) is inferior to that of the first group (Intermediate).
This indicates that it is more difficult  to classify cell images when their contrast is low.
However, the improvement achieved by mutual information regularization is consistent over these two groups.

\section{Conclusions}
\label{sec:con}

Can knowing the classification response of a data sample reduce uncertainty about its true class label?
In this paper, we proposed this question and tried to answer it by learning an optimal classifier to reduce such uncertainty.
Insighted by the fact that the reduced uncertainty can be measured by the mutual information between classification responses and true class labels, we proposed a new classifier learning algorithm, by maximizing the mutual information when learning the classifier. Particularly, our algorithm adds a maximum mutual information regularization term. We investigated the classification performances when
maximum mutual information was used to regularize the classifier learning based on four different loss functions. The the experimental results show that
the proposed regularization can improve the classification performances of all these four loss function based classifiers.
In the future, we will study how to apply the proposed algorithm on large scale dataset based on some distributed big data platforms \cite{su2013sdquery,wang2013supporting,wang2012scimate,qingquan2010context} and use it to signal and power integrity applications \cite{liu2011frequency,wang2012passivity,lei2010vector,wang2010mfti,wang2010peds,lim2010ultra,lei2010decade}.

\section*{Acknowledgements}

This work was supported by grants from King Abdullah University of Science and Technology (KAUST), Saudi Arabia.

\end{document}